# A framework for mining lifestyle profiles through multi-dimensional and high-order mobility feature clustering[*]

YESHUO SHU

College of Architecture and Landscape, Peking University

Email: shuyeshuo@pku.edu.cn

GANGCHENG ZHANG

College of Architecture and Landscape, Peking University

Email: zgc1996@stu.pku.edu.cn

KEYI LIU

College of Architecture and Landscape, Peking University

Email: ky.liu@stu.pku.edu.cn

JINTONG TANG

College of Architecture and Landscape, Peking University

Email: wanzhoutjt@pku.edu.cn

LIYAN XU*

College of Architecture and Landscape, Peking University

Email: xuliyan@pku.edu.cn

Human mobility demonstrates a high degree of regularity, which facilitates the discovery of lifestyle profiles. Existing research has yet to fully utilize the regularities embedded in high-order features extracted from human mobility records in such profiling. This study proposes a progressive feature extraction strategy that mines high-order mobility features from users' moving trajectory records from the spatial, temporal, and semantic dimensions. Specific features are extracted such as travel motifs, rhythms decomposed by discrete Fourier transform (DFT) of mobility time series, and vectorized place semantics by word2vec, respectively to the three dimensions, and they are further clustered to reveal the users' lifestyle characteristics. An experiment using a trajectory dataset of over 500k users in Shenzhen, China yields seven user clusters with different lifestyle profiles that can be well interpreted by common sense. The results suggest the possibility of fine-grained user profiling through cross-order trajectory feature engineering and clustering.

---

[*] This is a working paper. For the preprint version, please see：

## 1 INTRODUCTION

Trajectory data mining offers insight into meaningful clusters and patterns within datasets. In particular, mining user lifestyle profiles reflected in trajectories is appealing in various domains, such as recommendation systems, travel planning, and social surveys. Especially in the context of increasing emphasis on data ethics, it is valuable to capture the lifestyle knowledge implicit in user trajectories while respecting anonymity. However, the common and well-studied research in trajectory clustering, whether for basic spatiotemporal trajectories, trajectories with uncertainties, or semantically enriched trajectories, focuses on discovering the generalized geometric similarities inherent in the trajectories themselves [3, 4, 25, 26]. Such similarities do not logically imply similarities in the users' lifestyles. Indeed, while there is a strong correlation between mobility patterns and demographics [17, 22], the patterns are abstract in nature and do not necessarily lead to geometric similarity. For example, consider two commuters who live and work in different parts of the city. While their lifestyles may be strikingly similar, their mobility trajectories may not match by any generalized geometric measure. Thus, even though distance-based clustering can generally still be the methodological foundation for mining lifestyle profiles, the feature space should be more abstract and reflect the essence of the "lifestyle" rather than a direct geometric measure of the trajectory.

The question arises: what trajectory features can efficiently depict an individual's lifestyle profile? An abundance of research has already extracted trajectory features from multiple dimensions, including spatial, temporal, and semantic aspects (listed in Table 1). Existing methods still have noticeable limitations: While a significant body of research has focused on clustering semantically enriched trajectories in the geometric-similarity-based mining approach [19, 25], attempts to integrate features from all three dimensions in lifestyle-similarity-based user profiling remain relatively preliminary. On the other hand, the extraction of trajectory features is often to some extent arbitrary. In particular, commonly used features focus on basic spatial, temporal, and semantic measurements or descriptions of the trajectories while lacking reference to more abstract features. Besides, the profiling results generally yield a coarse-grained user clustering, which could intuitively be achieved only by using such abstract features, leading to underutilizing the more specific mobility features.

Table 1: Common trajectory features organized by dimension and metric. Some examples are provided

| Dimension | Metric | Examples | Research |
|---|---|---|---|
| Space | Scale | number of stays, total distance | [17, 21] |
| | Distribution | centroid, radius of gyration | [18, 21] |
| | Geometry | ellipsoid of inertia, convex hull | [6, 7] |
| | Graph | motif | [9, 18] |
| Time | Scale | active days | [17] |
| | Distribution | average record time, fragmented degree | [21, 23] |
| | Periodicity | spectral features | [13, 24] |
| Semantic | Scale | number of visited locations | [1] |
| | Distribution | land use proportions | [21] |
| | Context | word2vec vector | [5] |

The root of the challenges above is the lack of theoretical support in user profiling. Statistical physical studies of human mobility have provided a solid foundation for understanding trajectory characteristics as a reflection of individual lifestyles. First, the dichotomy of human mobility patterns as "explorers" and "returners" allows for a coarse-grained categorization of mobility-defined human groups. Second, by examining the trajectory's spatial range [7] and temporal rhythm [20], we can further differentiate lifestyle variations influenced by gender, age, and occupation. Finally, direct descriptions of trajectory features such as location [11] and place semantics [5] offer specific characterizations of personalized identity. Overall, the above theoretical framework hints at a progressive profiling process that moves from abstraction to specificity, in which coarse-grained lifestyle groups are distinguished by "high-order" mobility features or "feature of features" such as repetitiveness of movement patterns or activity ranges. Fine-grained personal identities are revealed using more straightforward descriptions of geometric, temporal, and semantic trajectory features, namely "first-order" mobility features.

Based on the above ideas, this paper proposes a methodological framework to discover users' lifestyle profiles. The framework builds on a widely adopted trajectory mining procedure, which includes two steps: feature extraction and clustering [21]. The first step is feature engineering, where we distinguish between "first-order" and "higher-order" features: the former encapsulates the basic geometric, temporal, and semantic attributes of the trajectories, while the latter exploits the statistical, relational, and pattern information stemming from the first-order features. In the second step, the extracted features are clustered. Due to the anonymity of trajectory records, supervised classification or end-to-end deep learning methods are usually not applicable in such clustering. Instead, researchers often use unsupervised clustering methods to identify groups of similar trajectories within the feature space [17]. The framework has two bright sides. First, the clustering results are expected to be interpreted coherently based on theoretical foundations. Second, multi-dimensional first-order features are retained, which can be flexibly selected in fine-grained analysis to address different scenarios and tasks (Figure 1).

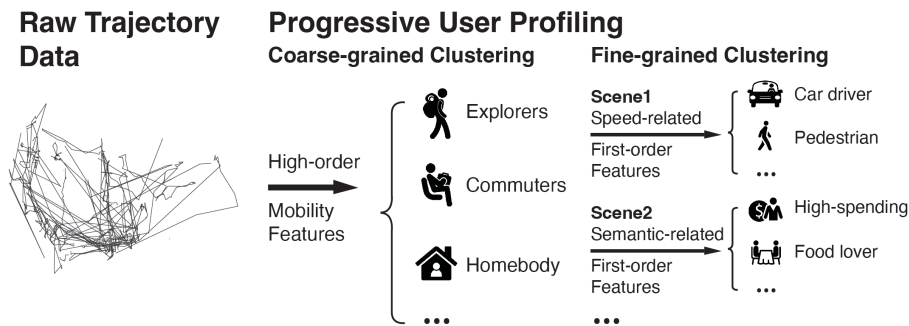

Figure 1: A representation of the proposed progressive user profiling framework

## 2 METHODOLOGYS

### 2.1 Definitions

**Trajectory**: A trajectory is a sequence of $(lng, lat, t)$ tuples recorded chronologically by a user's mobile device during a given period, where $lng$ and $lat$ represent the coordinates approximated to the center of a $l$-meter grid after anonymization, and $t$ means the timestamp. In the dataset, $l$ is set to 150.

**Stay**: If a grid is recorded continuously for more than $T$ minutes, it is considered a stay and is denoted by $(G_{id}, t)$ where $T$ is the time threshold, $G_{id}$ is the grid id, and $t$ is the duration of the stay. In this study, $T$ is set to 30.

**Stay List**: A stay list is the user's stays in chronological order and is denoted as $[stay_1, stay_2, \ldots, stay_j]$, where $j$ is the total number of stays of this user.

**Motif**: A motif is a directed graph structure derived from a user's stay list. In this graph, each stay is a node, and the trips between stays are edges.

**Mobility Rhythm**: Divide the 24-hour day into time bins of equal length and calculate the proportion of the travel distance in each bin to the total distance. The resulting sequence is called the mobility rhythm and is denoted by $[d_1, d_2, \ldots, d_k]$, where $k$ is the number of bins. In this study, $k$ is set to 12.

**Activity Semantic**: Each user's stay is matched with Point of Interest (POI) or Area of Interest (AOI) data based on spatial proximity. This match generates the activity semantic, denoted as $[POI_{tag_1}, POI_{tag_2}, \ldots, POI_{tag_j}]$, where $POI_{tag_i}$ represents the semantic tag of the matching POI/AOI for the $i$-th stay, and $j$ represents the length of the user's stay list.

## 2.2 Feature Framework and Clustering

Following are the proposed high-order mobility features, which are also listed in Table 1:

High-order spatial features: The motif-based approach allows the processing of user trajectories in a directed graph structure, preserving the structural information of daily mobility patterns [18] Additionally, the radius of gyration, a concept borrowed from classical mechanics, characterizes the spatial extent of the user's activities [7].

High-order temporal features: Spectral analysis techniques, including Discrete Fourier Transform (DFT) and Wavelet Transform (DWT) [16, 24], can extract periodic patterns from time series of mobility features, such as travel distance and speed.

High-order semantic features: Researchers have found that trajectory nodes can be analogized to words in natural language [10], enabling the application of natural language models such as word2vec [5] and Latent Dirichlet Allocation (LDA) [8] to extract contextual relationships between travel activities.

### 2.2.1 Spatiotemporal features.

**Motif type (MT):** Research on human mobility has shown that a small number (<10) of motifs can effectively summarize the majority of trajectories [18]. In our dataset, we found that the top four motif categories had the highest proportion, covering 86.2% of users (see Figure 2). Therefore, we categorized the remaining motif types as "MT others" and performed one-hot coding on these five motif categories.

**Radius of gyration (ROG):** The radius of gyration of each individual is defined in kilometers and calculated using Equation 1

$$R_u = \sqrt{\frac{1}{n_u} \sum_{i=1}^{n_u} \left( r_i(u) - r_c(u) \right)^2} \qquad (1)$$

Where $r_i(u)$ denotes the location of the $i$-th trajectory point for user $u$, $r_c(u)$ denotes the centroid of all trajectory points for user $u$, and $n_u$ denotes the total number of trajectory points for user $u$. The higher the ROG, the greater the spatial activity range of an individual.

In the temporal dimension, we extracted the low-frequency information of the mobility rhythm using DFT. Then, features were constructed mainly focusing on the first two frequency domain coefficients corresponding to one cycle per day and two cycles per day. The choice of these low-order frequencies is based on two main reasons. First, the energy of the time series tends to be concentrated in the lower-order coefficients [13].

Second, the daily routines of the majority of individuals exhibit these two periodic patterns (as shown in Figure 2).

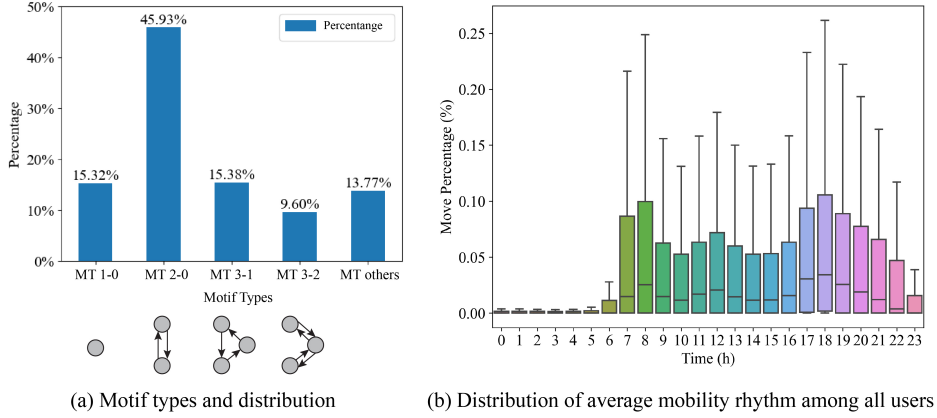

(a) Motif types and distribution

(b) Distribution of average mobility rhythm among all users

Figure 2: The distribution of users' motifs and average mobility rhythm

**Low frequency energy ratio (LFER)**: This feature quantifies the strength of low frequency periodicity in the user's mobility rhythm and is calculated using Equation 2

$$LFER = \frac{amp_1^2 + amp_2^2}{\sum_{i=1}^{k} amp_i^2} \qquad (2)$$

where $amp_i$ is the amplitude of the signal of period $i$ of the DFT-decomposed mobility rhythm, and $k$ is the length of the mobility rhythm. The higher the LFER, the more time-periodic the individual's mobility.

**Diurnal cycle frequency ratio (DCFR):** This feature quantifies the tendency of low frequency periodicity in a user's mobility rhythm and is calculated using Equation 3. A higher value indicates a higher probability of a two-cycle-per-day pattern, while a lower value indicates a higher probability of a one-cycle-per-day pattern.

$$DCFR = \frac{amp_2}{amp_1 + amp_2} \qquad (3)$$

### 2.2.2 *Semantic features.*

We used the word2vec model based on the Continuous Bag-of-Word (CBOW) word embedding algorithm to convert each node within the user activity semantics into a vector representation [12]. The model has two hyperparameters: $dim$, which represents the length of the embedding vector, and $w$, which is the context length. Based on previous research [5], we set these parameters to 80 and 2. Consequently, the user's activity semantic was transformed into a list $[v_1, v_2, \ldots, v_j]$, where $j$ represents the length of the user's activity semantic and $v_i$ represents the vectorized representation of the $i$-th activity semantic derived from the word2vec model. In the following sections, $v_i$ will be referred to as the semantic vector, while the list $[v_1, v_2, \ldots, v_j]$ will be referred to as the semantic list. We then developed three features:

**Unique activity semantic number ($N_{uas}$)**: This feature quantifies the richness of the user's semantic activities. It refers to the number of unique semantic vectors in the user's semantic list.

**Mean activity semantic ($M_{as}$)**: This feature captures users' daily activity context logic similarity. It is the average vector of all semantic vectors in the user's semantic list.

**Max semantic distance ($M_{sd}$)**: This feature captures the variability of the user's semantic activity. It is calculated by taking the maximum distance between any two different semantic vectors within the user's semantic list.

### 2.2.3 Clustering Method.

Since activity semantics can be analogized to natural language corpora, they require high-dimensional vectors for representation. As a result, the semantic feature dimension is not balanced with the spatiotemporal feature dimension. This, in turn, could lead to clustering within the feature space being predominantly driven by semantic information. In such cases, concatenating all features may not yield optimal clustering results. To address this challenge, we adopted a multi-view k-means clustering method [2, 15], which adapts the traditional k-means through a co-training process. The co-training process involves alternating between training algorithms, leveraging prior information or knowledge from each view to enhance consistency across different views. Specifically, this method is based on a co-EM algorithm; each view refines the model parameters during each iteration to maximize the likelihood based on the expected values of the hidden variables from the alternate view [2].

## 3 EXPERIMENTS

### 3.1 Data and Pre-processing

We used an anonymized location dataset provided by a Chinese Internet company, collected from users' mobile devices with informed consent when using the company's navigation application. The trajectory dataset consists of coordinates tuples (latitude and longitude), timestamps, and distinctive user identifiers. These user IDs serve solely for the purpose of distinguishing trajectories, and they do not provide any detailed user-specific information. To further protect users' privacy, the dataset underwent an anonymization process by the data provider in which precise location information was converted to the center coordinates of a 150-meter grid cell. All location information used in this study is derived from this grid-based processing.

The original data underwent several pre-processing steps described below: First, trajectory points with localization drift were eliminated to ensure data reliability. Then, records with abnormally high speeds and insufficient total recording duration were filtered out at 1% to ensure data quality.

The dataset consists of trajectory data from 501,158 users (after pre-processing) in Shenzhen on the weekday of December 4, 2019. On average, each user has approximately 125 trajectory points per day. Users tend to visit a limited number of locations in a single day; more than 75% of users have fewer than six unique stays. More than 90% of users have trajectory entries in more than ten different hours, resulting in a record length of more than 15 hours daily. A significant proportion, more than three-quarters of users, tend to travel short distances, averaging less than 20 km per day, with a total distance averaging 15.99 km.

The POI and AOI data are sourced from Baidu Maps, primarily comprising names and tags. Baidu classifies POI/AOI based on the standard of industry affiliations (https://lbs.baidu.com/index.php?title=open/poitags).

## 3.2 Result Analysis

The Spearman's correlation graph shows a relatively weak correlation between the features across the different views (see Figure 3). Seven user clusters are identified, revealing a hierarchical structure reminiscent of a tree diagram (see Figure 4). This hierarchical structure has four levels, with each level separating users based on different features:

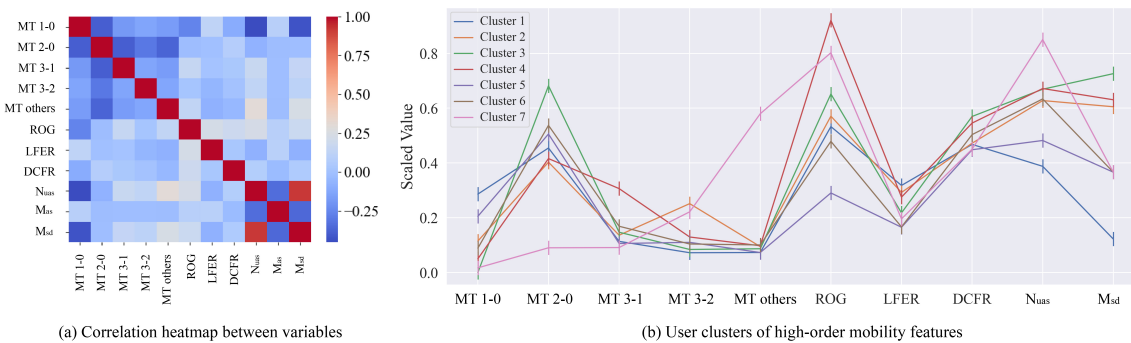

(a) Correlation heatmap between variables      (b) User clusters of high-order mobility features

Figure 3: Feature correlation and cluster differences

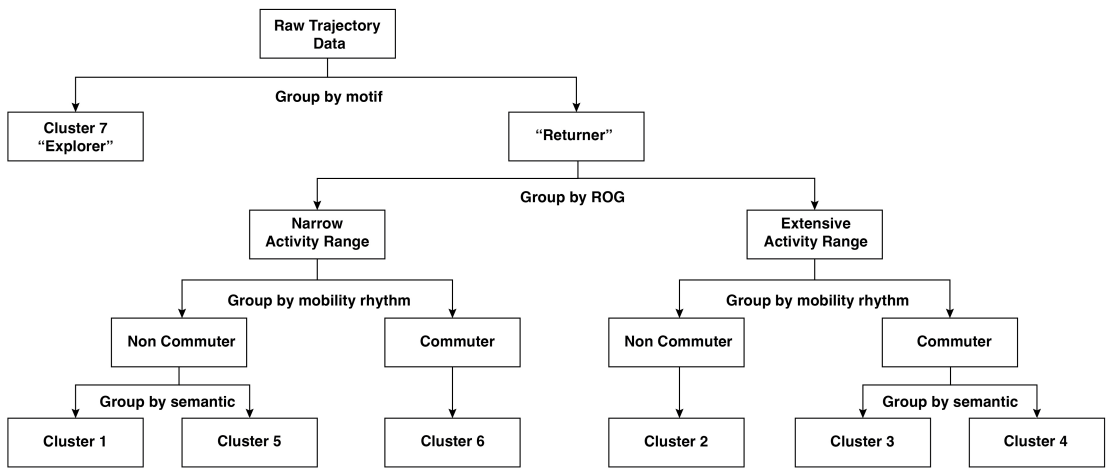

Figure 4: Hierarchical interpretation of the multi-view k-means clustering results

At the first level, the users were divided into two groups based on the motif's complexity: Use cluster 7 stands out, characterized by complex travel patterns, a wide range of activities, and rich semantics. They tend to be exploratory and recreational travelers. This cluster represents a small proportion, accounting for approximately 11% of all users in our dataset. This finding is consistent with the proportions of explorers in previous human mobility studies [14].

At the second level, users are divided into two groups based on ROG: large activity range and small activity range. Individuals with a large activity range are more likely to choose motorized vehicles or public transportation as their primary means of transportation, while those with a small activity range tend to prefer walking. It is noteworthy that user cluster 5 has a smaller activity range than all other categories, indicating a higher tendency to localize activities close to home.

At the third level, users are further divided into two groups based on the spectral features of the mobility rhythm: those likely to have two cycles and those unlikely to have two cycles. This distinction is thought to reflect the different rhythms of urban commuters and non-commuters. For example, individuals in user clusters 3 and 4 have high low-frequency energy and a significant two-cycle-per-day pattern, indicating that they are more likely to be regular commuters. In conjunction with the previous level, it is also possible to distinguish commuters with longer journeys from those with shorter journeys.

At the final level, the population is divided based on activity semantic similarity and richness. Within these identified groups, individuals exhibit regular activities and multiple visitation points. These individuals tend to have fixed daily destinations, such as parents responsible for getting their children to and from school.

To further validate the clustering results, we employed the Latent Dirichlet Allocation (LDA) model to identify prominent topics within the activity semantics of the seven user clusters (**Error! Reference source not found.**). These results were utilized to confirm and support our findings. For instance, the dominant topic associated with user cluster 1 primarily centers around the residence, weighing 0.826. This result suggests that their daily activities mainly revolve around their residential environment, consistent with our initial assumptions. In addition, user clusters 3 and 4 exhibit significant topical relevance to residence and company, indicating a high likelihood of being daily commuters. Lastly, user cluster 7 shows low weights assigned to the first four topics, totaling up to 0.652, which suggests a propensity for diverse modes of travel.

Table 1: Topics detected by the LDA model for user clusters. The third column is the sum of the top 4 topics' weights.

| Clusters | The four most significant topics | Aggregate weights |
|---|---|---|
| Cluster 1 | Residence (0.826), Internal building (0.068), Recreational plaza (0.029), Commercial space (0.015) | 0.938 |
| Cluster 2 | High school (0.455), Residence (0.357), Company (0.032), Commercial space (0.019) | 0.863 |
| Cluster 3 | Residence (0.502), Company (0.309), Commercial space (0.021), Chinese restaurant (0.015) | 0.847 |
| Cluster 4 | Residence (0.439), Office building (0.389), Company (0.035), Commercial space (0.028) | 0.891 |
| Cluster 5 | Company (0.682), Quick service restaurant (0.055), Residence (0.046), Convenience store (0.038) | 0.821 |
| Cluster 6 | Industrial park (0.495), Residence (0.257), Company (0.091), Office building (0.019) | 0.862 |
| Cluster 7 | Residence (0.389), Company (0.125), Park (0.088), Office building (0.05) | 0.652 |

In addition, the LDA results further enhance our understanding and interpretation of the clustering results: User cluster 2 is dominated by the "MT 2-0" and "MT 3-2" motifs, which correspond to the travel patterns of parents and students associated with the high school education topic. User cluster 5 has a topic of company and a small range of activities so that they may live and work nearby. User cluster 6 has a topic of the industrial park, which explains why they are likely to commute, but their range of activities is relatively small.

## 4 CONCLUSION AND FUTURE WORK

The cornerstone of this study is the construction of a refined feature framework with an emphasis on high-order features across spatiotemporal and semantic dimensions. The experimental results showed that a set of high-order features derived from trajectory data can effectively represent the characteristics of individuals and provide practical insights into their travel patterns and lifestyles. The interpretable clustering results supported the development of a progressive user profiling framework based on the differentiation between high-order and first-order features.

This study successfully identified seven distinct and easily identifiable individual mobility patterns using high-order features for coarse-grained clustering. We also plan to conduct further tests on high-precision GPS datasets with user labels to evaluate the effectiveness of our framework. Future work can focus on integrating more specific, targeted, and first-order mobility features for fine-grained clustering. However, this cross-order mining approach should have its inherent limitations. The delicate balance between trajectory data anonymity and data mining is worthy of profound contemplation.

By aggregating trajectory information into specific spatial units, we can gain a deeper understanding of the spatial structure of activities within different social groups (see Figure 5). This includes specific examples such as the spatial distribution of people's stay points and spatial interaction patterns derived from the trajectory context. It's noteworthy that this spatial structure is multifaceted; it not only changes dynamically over time, but also varies with different modes of transportation. Therefore, comparing this type of spatial structure with traditional and static urban spatial structures - such as the distribution of socioeconomic and natural attributes and the topological structure of road networks - can reveal the interplay between people and their environment.

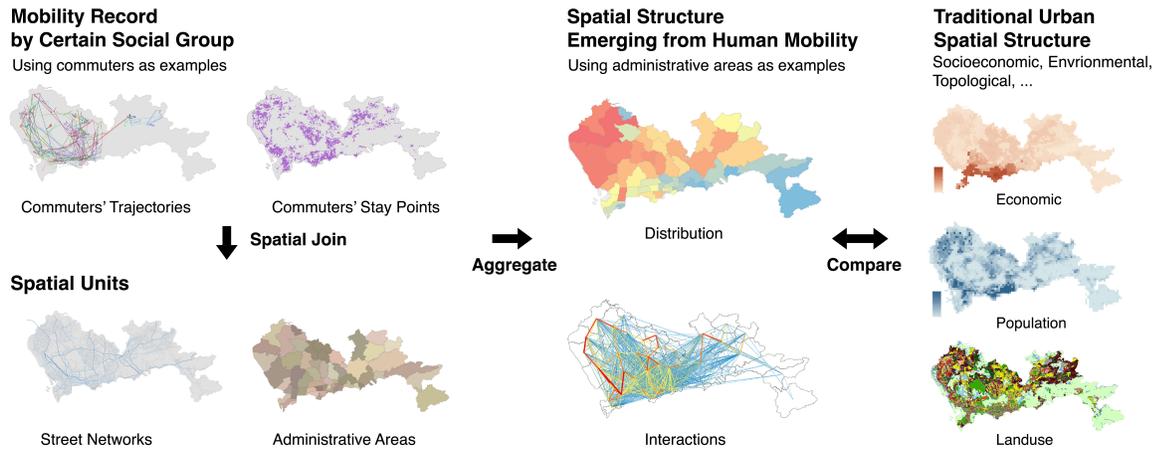

Figure 5: Illustration for comparisons between spatial structure of human mobility and traditional urban spatial structure